\documentclass[journal]{IEEEtran}
\usepackage[numbers,sort&compress]{natbib}
\usepackage{amsmath}
\usepackage{graphicx}
\usepackage{booktabs}
\usepackage{multirow}
\usepackage{xcolor}
\usepackage{url}
\usepackage[hidelinks]{hyperref}
\hypersetup{pdftitle={Incongruence-Aware Evaluation of Multimodal Speech Emotion Recognition in Audio Large Language Models},
  pdfauthor={Fengji Jing and Guy Laban}}
\usepackage{listings}
\newcommand{\pin}{\texorpdfstring{\ensuremath{p_{\text{incon}}}}{p\_incon}}
\begin{document}
\title{Incongruence-Aware Evaluation of Multimodal Speech Emotion Recognition in
Audio Large Language Models}
\author{Fengji~Jing$^{1}$ \quad and \quad Guy~Laban$^{2,3,4,*}$\\[6pt]
{\normalsize\itshape
$^{1}$Faculty of Science, The University of Hong Kong, Hong Kong, China\\
$^{2}$Department of Industrial Engineering and Management, Ben-Gurion University of the Negev, Beer Sheva, Israel\\
$^{3}$School of Brain Sciences and Cognition, Ben-Gurion University of the Negev, Beer Sheva, Israel\\
$^{4}$Azrieli National Center for Autism and Neurodevelopment Research, Ben-Gurion University of the Negev, Beer Sheva, Israel}\\[4pt]
{\normalsize $^{*}$Corresponding author: laban@bgu.ac.il}}
\markboth{Preprint}%
{Jing and Laban: Incongruence-Aware Evaluation of Multimodal SER in Audio LLMs}
\IEEEtitleabstractindextext{%
\begin{abstract}
Voice-interactive large language models process both what a speaker says and how
they say it, yet the two channels do not always agree. We ask whether that
disagreement degrades speech emotion recognition. Mapping text and audio into a
shared valence-arousal-dominance space, we convert the distance between them into a
continuous, threshold-free incongruence score with a two-component Gaussian mixture,
and validate the score against human judgement. We evaluate three audio-capable
models on acted, conversational and spontaneous speech (eNTERFACE'05, MELD and
CMU-MOSEI). With the model held constant across modalities, incongruence raises the
odds of error whenever audio is present, in every corpus, after controlling for the
ground-truth emotion and for distance in the feature space. Supervised classifiers
trained on the same features behave in the opposite way, which suggests that the
difficulty lies in how these models combine raw evidence rather than in the signal
itself. Text-only effects are unreliable: swapping only the model that reads the
transcript reverses their sign on one corpus. Rebuilding the score without the
emotion lexicon reproduces the central result, and among six simpler alternatives it
is the only label-free score that predicts error in all three corpora. Code, data
tables and prompts are released.
\end{abstract}
}
\maketitle
\IEEEdisplaynontitleabstractindextext
\IEEEpeerreviewmaketitle

\section{Introduction}
\label{sec:intro}

Voice-interactive large language models (LLMs) can simultaneously process the
acoustic signal of an utterance and its verbal content
\cite{chu2024qwen2audio,zeng2024glm4voice}, giving them access to both what a
user says and how they say it. This dual-channel capacity raises an immediate
and underexplored question: what happens when the two channels disagree? In
affective interactions \cite{Spitale2025PastWell-being}, inferring a user's
emotional state is already challenging from text alone
\cite{Mehra2025HowValues}; when speech and semantics are available together the
promise is richer understanding, but so is the risk of systematic confusion when
the channels conflict. The difficulty is compounded by a fundamental property of
natural speech: the acoustic signal and the verbal content are not always
aligned \cite{BanseScherer1996AcousticProfiles,Scherer2003VocalCommunication}.
Conflicting emotional signals in multimodal perception lead to systematic biases
in human observers, producing misunderstandings between \textit{what} a speaker
says and \textit{how} they say it
\cite{deGelder2000,Massaro1996,fantini2023consistency}. Even for human
listeners, judging whether an utterance's verbal and acoustic emotions conflict
is genuinely difficult \cite{Muller2011IncongruenceIntegration,
Ben-David2016ProsodySpeech}. As users increasingly share their emotions with
artificial agents \cite{laban2024sharing}, these agents must understand
disclosures even when emotional cues are incongruent.

Our goal is to \emph{evaluate}, rather than solve, cross-modal emotional
incongruence: to measure whether it degrades model performance, and to attribute
any degradation to cross-modal mismatch as an independent factor, separate from
the noise inherent in any affective annotation task. To do so we adopt the
Valence--Arousal--Dominance (VAD) framework \cite{bradley1994measuring,
pan2023review}, which encodes emotion as a continuous three-dimensional vector
and provides a shared geometric space in which the emotional content of text and
audio can be compared directly. We extract text-level VAD from the NRC VAD
Lexicon \cite{mohammad2018nrcvad} and audio-level VAD from a fine-tuned wav2vec2-large
model \cite{wagner2023dawn}, and propose a GMM-based scoring method that converts
the cross-modal distance into a continuous incongruence probability \pin{},
validated against human perception.

Two questions sharpen this goal. First, is any incongruence effect specifically
about \emph{multimodal} integration, or could it simply reflect transcripts that are
hard to read? Second, does it generalise beyond a single acted corpus and a single
model? The first question needs care, because a text-only condition is often run with
a different model from the multimodal one, and doing so quietly confounds model with
modality. Holding the model fixed, we find that incongruence raises error when the
model hears the audio, in acted, conversational and spontaneous speech alike, whereas
its association with text-only error is inconsistent and changes sign when a
different model reads the same transcripts. The penalty is therefore best understood
as a cost of combining two channels that disagree, not as a property of the words.

\paragraph{Contributions}
\begin{itemize}
\item \textbf{A validated, reusable incongruence-scoring tool.} A GMM-based
method that converts cross-modal VAD distance into a continuous, threshold-free
probability \pin{}, validated against human annotation (Spearman $r_s=0.65$),
reproducible from a public lexicon for any dual-modality speech dataset, and compared
with six simpler stratifiers, over which its advantage is consistency across corpora
rather than strength in any one of them.
\item \textbf{A paradigm dissociation.} On eNTERFACE, across three audio LLMs and
twelve conditions, \pin{} significantly and independently predicts LLM
classification error, whereas \emph{supervised} discriminative baselines trained on
the same VAD features show the opposite pattern (incongruence, if anything, helps
them). This is consistent with the hypothesis that the effect arises during raw
multimodal evidence integration rather than from VAD feature geometry alone, though
we cannot fully separate it from the supervised-versus-prompted contrast between the
two model classes.
\item \textbf{Evidence that the cost is multimodal.} Across acted, conversational
and spontaneous corpora, and with the model held constant, incongruence raises the
odds of error whenever audio is present. A controlled ablation shows that text-only
associations are an unreliable guide: swapping only the model that reads the
transcript reverses their sign on one corpus.
\item \textbf{An incongruence-aware benchmark.} Per-sample VAD features, \pin{}
scores, validity-checked LLM predictions, prompts, and analysis code, released
openly.
\end{itemize}

\section{Related Work}
\label{sec:related}

Our study sits at the intersection of four lines of research: multimodal and
speech emotion recognition, the psycholinguistics and computation of cross-modal
emotional incongruence, large audio-language models and their evaluation, and the
dimensional (VAD) account of affect. We review each in turn and position our
contributions at the close of the relevant subsections.

\subsection{Multimodal and speech emotion recognition}
\label{sec:rw-ser}
Emotion is encoded redundantly and complementarily across the lexical content of
an utterance, its vocal prosody, and (when available) the face, and recognising
it has long been treated as a multimodal fusion problem
\citep{baltrusaitis2019mml,pan2023review,wang2025mmersurvey}. Progress has been
driven by benchmark corpora that differ sharply in how emotion is elicited and
labelled: acted dyadic interaction (IEMOCAP; \citealp{busso2008iemocap}),
validated audio-visual acted speech and song (RAVDESS;
\citealp{livingstone2018ravdess}), induced reactions to scripted scenarios
(eNTERFACE'05; \citealp{Martin2006TheDatabase}), naturalistic podcast speech
retrieved at scale (MSP-Podcast; \citealp{lotfian2019msppodcast}), spontaneous
opinion monologues annotated for sentiment and emotion (CMU-MOSEI;
\citealp{zadeh2018mosei}), and multi-party conversational dialogue from
television (MELD; \citealp{poria2019meld}). These corpora vary not only in
spontaneity but also in how transparently the words alone reveal the speaker's
emotion, a property that prior work has rarely measured.

A parallel methodological divide concerns the emotion representation itself.
Categorical schemes assign utterances to discrete classes following the
basic-emotion tradition \citep{ekman1992basic,ekman1971constants}, whereas
dimensional schemes locate affect in a continuous space, most commonly
valence-arousal-dominance \citep{russell1980circumplex,mehrabian1974approach}.
Fusion strategies range from early concatenation through late decision-level
combination to attention-based intermediate fusion
\citep{atrey2010fusion,baltrusaitis2019mml}, and modern speech emotion
recognition (SER) is dominated by self-supervised representation learning that has
largely displaced hand-crafted features
\citep{schuller2018ser,latif2023survey,mustaqeem2024mser}, with benchmarks such
as MuSe pushing continuous in-the-wild affect prediction \citep{stappen2021muse}.
A standing assumption across this literature is that adding modalities can only
help; our results complicate this, showing that on identical VAD features the very
signal (cross-modal conflict) that degrades a generative model can \emph{improve}
a discriminative one.

\subsection{Cross-modal emotional incongruence}
\label{sec:rw-incongruence}
That prosody and lexical-semantic meaning are dissociable but interacting channels
is well established in psycholinguistics. Behavioural and neurophysiological
studies show that listeners integrate emotional voice and emotional words
automatically and cannot fully suppress one channel when attending to the other,
with incongruent pairings incurring measurable processing costs
\citep{deGelder2000,Massaro1996,Kotz2007,Muller2011IncongruenceIntegration}.
Using the Test for Rating of Emotions in Speech, \citet{Ben-David2016ProsodySpeech}
show that the two channels are separate but not separable and that prosody often
dominates semantics in the percept, a finding mirrored by vocal-expression
research on the acoustic profiles of emotion
\citep{BanseScherer1996AcousticProfiles,Scherer2003VocalCommunication}. This human
evidence motivates treating incongruence as a graded, perceptually grounded
quantity rather than a binary label.

Computationally, cross-modal conflict surfaces most visibly in sarcasm and irony,
where literal lexical sentiment is contradicted by prosodic or visual cues
\citep{Castro2019mustard}, and in multimodal sentiment analysis more broadly,
where contradictory modalities are a recognised failure mode
\citep{lai2023msasurvey}. \citet{wang2024inconsistency} construct an explicitly
modality-conflicting test set and report substantial degradation for both
classical fusion models and multimodal LLMs, while work on consistency and
incongruity in affective computing quantifies cross-channel agreement and its
effect on prediction \citep{fantini2023consistency,li2024incongruity}. What this
body of work largely lacks is a single, continuous incongruence score that is
computed in a shared interpretable affect space, validated against human
perception of mismatch, and used as a regressor to predict model error rather than
as a category to be classified. Supplying and validating such a score is our first
contribution.

\subsection{Audio-language models and their evaluation}
\label{sec:rw-alm}
The integration of speech and audio encoders with pre-trained LLMs has produced a
rapidly growing family of audio-language models capable of instruction following
over spoken input
\citep{zhang2023speechgpt,rubenstein2023audiopalm,chu2023qwenaudio,tang2024salmonn,chu2024qwen2audio}.
Recent systems move toward end-to-end spoken interaction, tokenising and
generating speech directly \citep{zeng2024glm4voice}, and
proprietary omni-modal models now accept audio natively alongside text
\citep{openai2024gpt4o}. These models inherit the transformer architecture and
large-scale pre-training that underpin text LLMs
\citep{vaswani2017attention,devlin2019bert,brown2020fewshot}, raising the question
of whether their emotional understanding is genuinely acoustic or merely lexical.

Evaluation has struggled to keep pace. General-purpose suites probe broad auditory
competence and instruction following
\citep{huang2024dynamicsuperb,yang2024airbench}, and recent surveys attempt to
organise this fragmented landscape \citep{yang2025lalmsurvey}. Most directly
relevant to us is a cluster of recent studies showing that audio-LLMs tend to
transcribe rather than listen: \citet{chen2026listen} find consistent lexical
dominance in SER, with models defaulting to neutral when words are uninformative
and failing under cue conflict; \citet{billa2026modality} show that under
controlled audio-text contradiction, audio-LLMs follow the text far more often
than they arbitrate toward a conflicting voice, even when instructed to trust it;
and acoustic-consistency suites penalise misalignment between semantic content and
tone \citep{maimon2025salmon}. Prompting and contextual framing also materially
affect SER accuracy in these models \citep{zhang2024contextualser}. We extend this
thread in two ways. First, rather than constructing adversarial cue conflicts, we
measure naturally occurring incongruence on a continuous scale and show that it
predicts LLM error across three corpora. Second, we show that how a text-only
baseline is built matters: when the transcript is read by a different model from the
one that hears the audio, model and modality are confounded, and the apparent
text-only effect of incongruence can reverse.

\subsection{Dimensional affect and VAD extraction}
\label{sec:rw-vad}
Our analysis depends on mapping both text and audio into a common
valence-arousal-dominance space. The dimensional view of affect descends from
circumplex and PAD models \citep{russell1980circumplex,mehrabian1974approach} and
is operationalised through self-report instruments such as the Self-Assessment
Manikin \citep{bradley1994sam,bradley1994measuring} and broader affective-computing
frameworks \citep{picard1997affective,calvo2010affect}. On the text side, large
human-rated lexicons provide per-word VAD norms, including the widely used NRC-VAD
lexicon \citep{mohammad2018nrcvad,Mohammad2025} and the Warriner norms
\citep{warriner2013norms}, enabling transcript-level affect estimation. On the
audio side, self-supervised models fine-tuned for dimensional emotion regress
valence, arousal, and dominance directly from the waveform \citep{wagner2023dawn}.
Casting both modalities into the same VAD coordinates lets us define a
modality-symmetric, interpretable distance and fit a Gaussian-mixture model whose
posterior yields our continuous incongruence score \pin{}. Building on this shared
space, our final contribution is a paradigm dissociation: on the same VAD
features, incongruence harms generative audio-LLMs while \emph{helping}
discriminative classifiers, which is consistent with the cost of cross-modal
conflict arising from how these models integrate evidence rather than from the
features themselves. We treat this as an interpretation consistent with the data,
not a direct probe of the models' internal integration mechanism.

\section{Method}
\label{sec:method}

\begin{figure*}[t]
  \centering
  \includegraphics[width=0.96\textwidth]{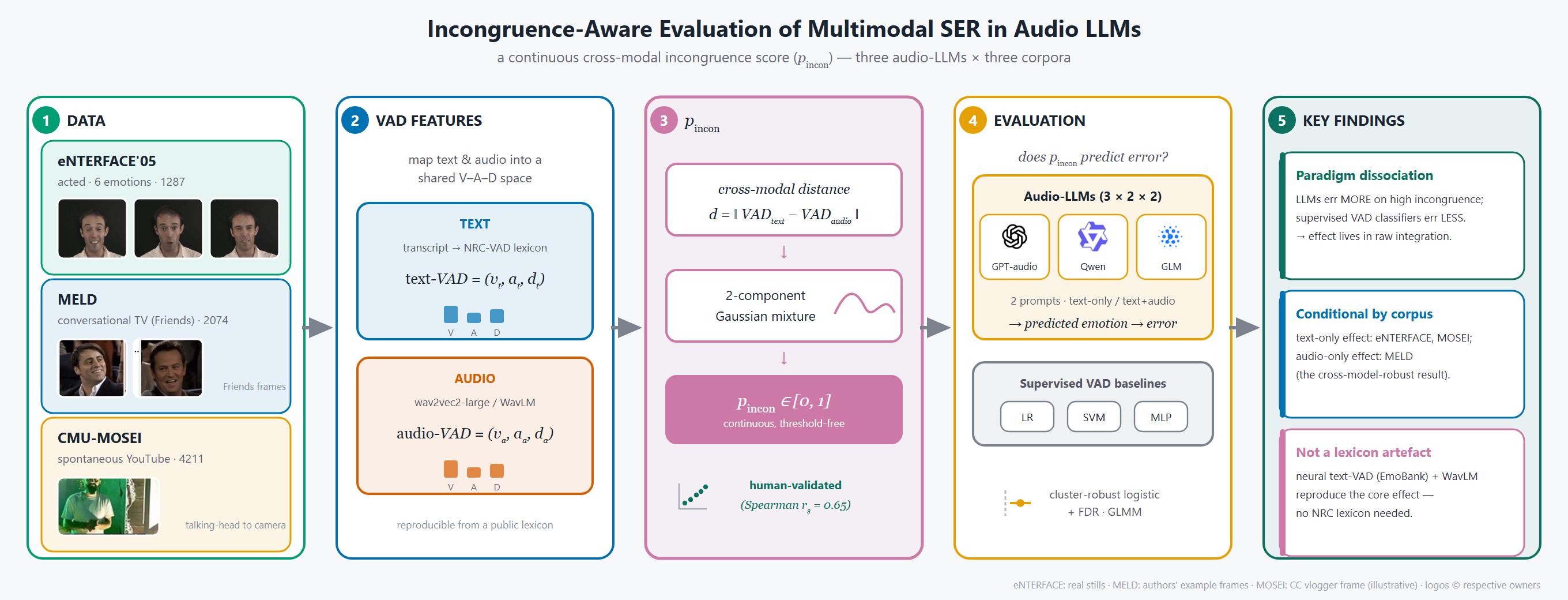}
  \caption{Overview. Text and audio are mapped into a shared VAD space; their Euclidean distance is converted to a continuous incongruence probability \pin{} via a two-component GMM fitted on training distances. LLM predictions are evaluated stratified by \pin{}, and compared against VAD-feature baselines.}
  \label{fig:overview}
\end{figure*}

Figure~\ref{fig:overview} summarises the pipeline. Both channels of each utterance
are mapped into a shared VAD space, their disagreement is turned into a continuous
incongruence score, and that score is then used to predict the errors of audio LLMs
and of classifiers trained on the same features.

\subsection{VAD feature extraction (unified across corpora)}
The same extractor is applied to every corpus, so results are directly
comparable and reproducible from public resources.

\paragraph{Text VAD}
Each transcript is lowercased and tokenised with NLTK's \texttt{word\_tokenize}.
For every token present in the NRC VAD Lexicon \cite{mohammad2018nrcvad} we retrieve
its (valence, arousal, dominance) triple on the bipolar $[-1,1]$ scale, and take
the element-wise mean over matched tokens as the text-level VAD vector.
Utterances whose tokens are entirely outside the lexicon (for example short
backchannels such as ``Okay.'' or ``What?'') have no defined text VAD and are
excluded; this is rare in read speech but, as we show, non-negligible in
conversational speech.

\begin{figure}[t]
  \centering
  \includegraphics[width=\columnwidth]{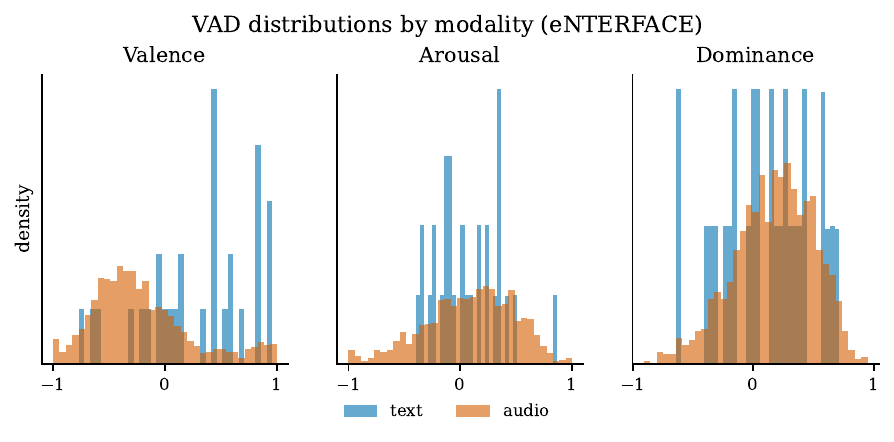}
  \caption{Distributions of the extracted valence, arousal, and dominance values
  by modality on eNTERFACE (text from the NRC-VAD lexicon, audio from the
  fine-tuned wav2vec2 regressor). The two modalities occupy overlapping but
  distinct regions of the VAD space, and their per-utterance discrepancy is what
  \pin{} quantifies.}
  \label{fig:vad}
\end{figure}

\paragraph{Audio VAD}
Audio-level VAD is extracted with a fine-tuned wav2vec2-large model trained on
MSP-Podcast for dimensional emotion regression \cite{wagner2023dawn}. Model
outputs (arousal, dominance, valence) are clipped to $[0,1]$ and linearly
rescaled to $[-1,1]$ via $x' = (x-0.5)\times 2$ to match the symmetric VAD
space. Figure~\ref{fig:vad} shows the resulting per-dimension distributions for
the two modalities.

\subsection{Quantifying incongruence: \pin{}}
The cross-modal distance for utterance $i$ is the Euclidean distance $d_i$
between its text and audio VAD vectors; larger $d_i$ indicates greater conflict.
To convert distance into a probabilistic score without an arbitrary threshold,
we fit a two-component Gaussian Mixture Model on the \emph{training-split}
distances (avoiding leakage):
\begin{equation}
  p(d) = \pi_1\,\mathcal{N}(d \mid \mu_1, \sigma_1^2)
       + \pi_2\,\mathcal{N}(d \mid \mu_2, \sigma_2^2).
\end{equation}
The component with the larger mean $k^*$ is the incongruent cluster, and the
incongruence probability is its posterior responsibility:
\begin{equation}
  p_{\text{incon}}(i) = p(z = k^* \mid d_i)
  = \frac{\pi_{k^*}\,\mathcal{N}(d_i \mid \mu_{k^*}, \sigma_{k^*}^2)}
         {\sum_j \pi_j\,\mathcal{N}(d_i \mid \mu_j, \sigma_j^2)} \in [0,1].
\end{equation}
We label utterances \emph{congruent} if $\pin{}<0.5$ and \emph{incongruent}
otherwise, but the continuous score is the primary predictor.

\paragraph{Lexical emotional transparency}
To characterise a corpus independently of any model, we measure how well each
modality alone identifies the emotion label. We build per-class VAD prototypes
(training-set means) and classify each utterance by nearest prototype using its
text VAD and, separately, its audio VAD. The resulting text-to-label and
audio-to-label accuracies quantify how much of the emotion is carried lexically
versus acoustically, and let us ask whether that shapes where incongruence harms
LLMs (Section~\ref{sec:crosscorpus}).

\subsection{LLM inference}
We evaluate three audio LLMs: \textbf{GPT-audio} (OpenAI), \textbf{Qwen2-Audio-7B-Instruct}
(Alibaba), and \textbf{GLM-4-Voice} (ZhipuAI). Decoding is greedy where
supported; models output a single JSON object \texttt{\{"label":"<emotion>"\}}
over the corpus label set, parsed by JSON extraction with a deterministic
keyword fallback. Responses failing both, or returning API errors, are excluded
via per-model validity flags. Two prompt templates are compared: \textbf{T1}, a
minimal zero-shot prompt giving only the label names and a JSON constraint; and
\textbf{T2}, a rule-guided prompt adding per-label reference cues and an explicit
conflict-resolution instruction. Two input modalities are compared:
\textit{text-only} and \textit{text+audio}. (An audio-only configuration was
piloted but discarded because GLM-4-Voice failed to produce parseable labels
under it.) Full prompt texts are included in the released repository.

\textbf{Reproducibility details.} For GPT-audio we use \texttt{gpt-audio-1.5} for
\emph{both} the text-only and the text+audio conditions (OpenAI Chat Completions,
accessed June~2026), so that modality is varied with the model held constant. Running
the text-only cells with a separate text model is a common shortcut, but it
confounds model with modality; we quantify the consequence by also running them with
\texttt{gpt-5.2}, a text-only model, as a controlled ablation
(Section~\ref{sec:modelabl}). The Qwen results on MELD/MOSEI use
\texttt{qwen-omni-turbo} (Qwen2.5-Omni, the audio-capable successor available on
our DashScope endpoint; the eNTERFACE Qwen results use Qwen2-Audio-7B-Instruct).
Full-corpus GLM-4-Voice results are reported for eNTERFACE; GLM-4-Voice is additionally evaluated on matched stratified samples from MELD and MOSEI in Table~\ref{tab:crossmodel}. Decoding uses greedy/default settings
with the model's native limits; audio is resampled to 16~kHz mono, encoded as
WAV, and clips longer than 30~s are truncated (affecting only MOSEI). Outputs are
parsed by JSON extraction with a deterministic keyword fallback; per-condition
invalid (unparseable or out-of-vocabulary) rates are reported in
Table~\ref{tab:invalid}. Invalid rates are low on eNTERFACE, rise to about
$5\%$ for MELD text+audio, and reach approximately $14$--$16\%$ on MOSEI in
both modalities; all invalid responses are excluded via per-condition
validity checks.

\begin{table}[t]
\caption{Invalid-response rate (\%) per condition (GPT-audio): responses that
were unparseable or out-of-vocabulary and hence excluded.}
\label{tab:invalid}
\centering\footnotesize
\resizebox{\ifdim\width>\columnwidth \columnwidth\else \width\fi}{!}{%
\begin{tabular}{lcccc}
\toprule
\textbf{Corpus} & \textbf{T1 txt} & \textbf{T1 t+a} & \textbf{T2 txt} & \textbf{T2 t+a} \\
\midrule
eNTERFACE & $0.0$ & $1.4$ & $0.5$ & $1.5$ \\
MELD      & $0.0$ & $5.5$ & $0.0$ & $4.9$ \\
MOSEI     & $15.8$ & $14.5$ & $13.6$ & $14.7$ \\
\bottomrule
\end{tabular}}
\end{table}

\subsection{Evaluation protocol and statistics}
Classification is assessed by accuracy and Macro-F1. To test the incongruence
effect we use three complementary analyses, all on valid predictions:
(i) \emph{group} comparison of congruent versus incongruent accuracy/Macro-F1;
(ii) a per-utterance logistic regression of error on \pin{} with text length and
audio duration as controls,
$\Pr(y_i{=}1)=\sigma(\alpha+\beta\,\pin{}_{,i}+\gamma_1\text{len}_i+\gamma_2\text{dur}_i)$,
where a positive, significant $\beta$ indicates that higher incongruence raises
error odds independently of trivial factors; and (iii) a $2\times2$
$\chi^2$ test (congruent/incongruent $\times$ correct/error) with odds ratio and
Cram\'er's $V$. Because congruent and incongruent subsets need not share the same
class distribution, we additionally report a \emph{within-emotion} control that
re-tests the effect inside each ground-truth class. These per-condition analyses
are descriptive. Our primary inferential test is a single cluster-robust logistic
model per corpus that pools the four conditions, includes a \pin{}$\times$audio
interaction and controls for emotion class and feature-space geometry
(Section~\ref{sec:rigor}).

\section{Datasets}
\label{sec:data}

\paragraph{eNTERFACE'05} An audio-visual corpus of $1{,}287$ scripted emotional
utterances over six emotions (anger, disgust, fear, happiness, sadness,
surprise) \cite{Martin2006TheDatabase}, split 8:1:1 into train/validation/test.
Sentences are written to express a target emotion, so the verbal content is
emotionally explicit.

\paragraph{MELD} A multi-party conversational corpus derived from the television
series \emph{Friends} \cite{poria2019meld}, with seven emotion labels (the six
above plus neutral). MELD is acted (professional actors performing scripts) but,
unlike eNTERFACE, its lines are conversational dialogue rather than
emotion-prototypical sentences, so emotion is carried largely by delivery and
context rather than word choice. We use the test split ($2{,}610$ utterances;
$2{,}074$ retain a defined text VAD).

\paragraph{CMU-MOSEI} A spontaneous, openly licensed corpus of unscripted
opinion monologues from online video \cite{zadeh2018mosei}, annotated for the
same six Ekman emotions. Per-utterance emotion intensities are reduced to a
single label by taking the dominant emotion (argmax). MOSEI adds genuinely
spontaneous speech to the two acted corpora: opinion monologues are unscripted yet
often lexically explicit (``I loved it'', ``it was terrible''). We process the
$4{,}211$ non-neutral utterances that retain a defined text VAD.

\section{Results}
\label{sec:results}

\subsection{eNTERFACE'05: incongruence is associated with LLM error}
Table~\ref{tab:enterface} reports all twelve conditions. GPT-audio leads
overall (up to $0.907$ accuracy under T2 text+audio). Descriptive high- versus
low-\pin{} differences vary across conditions. In the condition-wise
continuous regressions, Qwen2-Audio shows positive coefficients in all four
conditions, whereas GPT-audio is significant under T1 text+audio and both T2
conditions but not under T1 text-only. GLM-4-Voice is direction-unstable,
with a positive coefficient under T2 text-only and null or negative estimates
when audio is supplied. The pooled, cluster-robust analysis with the full
control set is reported in Table~\ref{tab:rigor} and is treated as the primary
inferential result. The two analyses disagree on the text-only cells. Per condition,
GPT-audio's text-only coefficients are positive or null, but once emotion class and
prototype distance enter the pooled model the text-only association turns
protective, whereas the text+audio association is positive under both. This
instability of the text-only estimate recurs throughout the paper and is one reason
we draw no conclusions from it.

\begin{table}[t]
\caption{eNTERFACE'05 (test split): overall accuracy and Macro-F1; congruent vs
incongruent accuracy; and the control-adjusted logistic effect of \pin{} on
error (all-split, with text-length and audio-duration controls).
$^{*}p<0.05$, $^{**}p<0.01$, $^{***}p<0.001$. OR is the unadjusted high-versus-low
incongruence odds ratio; $\beta$ is the coefficient from the adjusted continuous
logistic model.}
\label{tab:enterface}
\centering\footnotesize
\resizebox{\ifdim\width>\columnwidth \columnwidth\else \width\fi}{!}{%
\begin{tabular}{llcccccc}
\toprule
\textbf{Model} & \textbf{Cond.} & \textbf{Acc.} & \textbf{F1} &
\textbf{Acc$_c$} & \textbf{Acc$_i$} & \textbf{OR} & $\boldsymbol{\beta}$ \\
\midrule
\multirow{4}{*}{GLM-4-Voice}
 & T1 txt & .686 & .662 & .722 & .630 & 1.20 & $+.02$ \\
 & T1 t+a & .627 & .530 & .667 & .565 & 0.69 & $-.18^{**}$ \\
 & T2 txt & .729 & .727 & .764 & .674 & 1.36 & $+.14^{*}$ \\
 & T2 t+a & .720 & .709 & .750 & .674 & 0.81 & $-.14^{*}$ \\
\midrule
\multirow{4}{*}{Qwen2-Audio}
 & T1 txt & .767 & .751 & .767 & .766 & 1.35 & $+.18^{*}$ \\
 & T1 t+a & .700 & .657 & .726 & .660 & 1.27 & $+.15^{*}$ \\
 & T2 txt & .633 & .576 & .685 & .553 & 1.68 & $+.32^{***}$ \\
 & T2 t+a & .633 & .603 & .699 & .532 & 2.51 & $+.55^{***}$ \\
\midrule
\multirow{4}{*}{GPT-audio}
 & T1 txt & .847 & .844 & .873 & .809 & 1.30 & $+.03$ \\
 & T1 t+a & .873 & .871 & .901 & .830 & 1.77 & $+.20^{*}$ \\
 & T2 txt & .814 & .814 & .831 & .787 & 1.70 & $+.19^{*}$ \\
 & T2 t+a & .907 & .905 & .944 & .851 & 2.59 & $+.49^{***}$ \\
\bottomrule
\end{tabular}}
\end{table}

\subsection{Per-emotion vulnerability and a within-class control}
\label{sec:peremotion}
\begin{figure*}[t]
  \centering
  \includegraphics[width=0.78\linewidth]{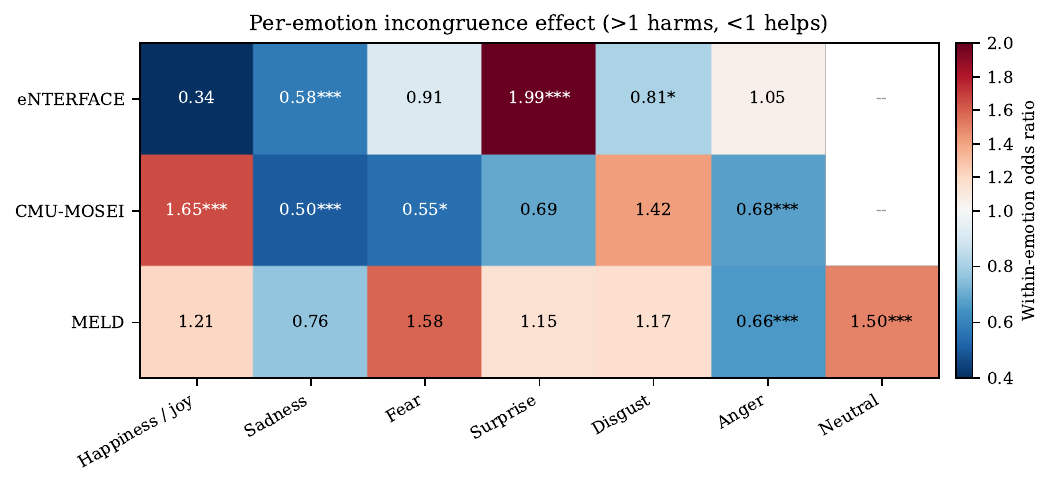}
  \caption{Within-emotion incongruence effect (OR of error) by corpus. Values
  ${>}1$ (red) mean incongruence harms recognition of that emotion; ${<}1$ (blue)
  mean it helps. Sadness and anger, the classes with the most characteristic prosody, are
  helped wherever their effect is significant; the classes that are harmed differ by
  corpus.
  (MELD has no ``happiness'' class; its positive label is ``joy''.)}
  \label{fig:heatmap}
\end{figure*}

The incongruence penalty is heterogeneous across emotions. A first caution is
that, on the eNTERFACE test split, the congruent and incongruent subsets do
\emph{not} share the same class distribution
(label $\times$ subset $\chi^2=18.2$, $p=.0027$): sadness and anger together
make up more than half of the incongruent subset, while surprise and happiness
each contribute only about a tenth. This means a naive subset comparison is partly confounded, and
that subset Macro-F1 can move more than accuracy. We therefore re-test the effect
\emph{within} each ground-truth emotion, where class composition is held fixed
(Table~\ref{tab:peremotion}, Fig.~\ref{fig:heatmap}). The per-utterance logistic regression in
Table~\ref{tab:enterface}, being item-level, is also immune to this confound and
remains the primary inferential test.

\begin{table}[t]
\caption{Within-emotion incongruence effect: odds ratio for misclassification of
high- versus low-\pin{} items, holding the ground-truth class fixed
(pooled over all splits and all 12 LLM conditions for eNTERFACE; GPT-audio
conditions for MELD and MOSEI). OR${}>1$ =
incongruence harms; OR${}<1$ = incongruence helps. $^{*}p<0.05$,
$^{***}p<0.001$; ns $=$ not significant.}
\label{tab:peremotion}
\centering\footnotesize
\resizebox{\ifdim\width>\columnwidth \columnwidth\else \width\fi}{!}{%
\begin{tabular}{lccc}
\toprule
\textbf{Emotion} & \textbf{eNTERFACE} & \textbf{MELD} & \textbf{MOSEI} \\
\midrule
surprise   & $1.99^{***}$ (harms) & $1.15$ ns            & $0.69$ ns \\
disgust    & $0.81^{*}$ (helps)   & $1.17$ ns            & $1.42$ ns \\
fear       & $0.91$ ns            & $1.58$ ns            & $0.55^{*}$ (helps) \\
anger      & $1.05$ ns            & $0.66^{***}$ (helps) & $0.68^{***}$ (helps) \\
sadness    & $0.58^{***}$ (helps) & $0.76$ ns            & $0.50^{***}$ (helps) \\
happiness / joy & $0.34$ ns       & $1.21$ ns            & $1.65^{***}$ (harms) \\
neutral    & --                    & $1.50^{***}$ (harms) & -- \\
\bottomrule
\end{tabular}}
\end{table}

The pattern differs by class and by corpus. Incongruence goes with more errors
for surprise on eNTERFACE, for neutral on MELD, and for happiness on MOSEI.
It goes with fewer errors for sadness and disgust on eNTERFACE, for anger on
MELD, and for fear, anger, and sadness on MOSEI. One regularity holds across
corpora: sadness and anger, the classes with the most characteristic prosody,
are helped wherever their effect is significant, whereas the classes that are
harmed vary, and on MELD and MOSEI they are the dominant label of the corpus.
Because these estimates pool prompts and modalities, they do not isolate an
audio-specific effect.

\subsection{Paradigm dissociation: LLMs versus VAD-feature classifiers}
\begin{figure}[t]
  \centering
  \includegraphics[width=\columnwidth]{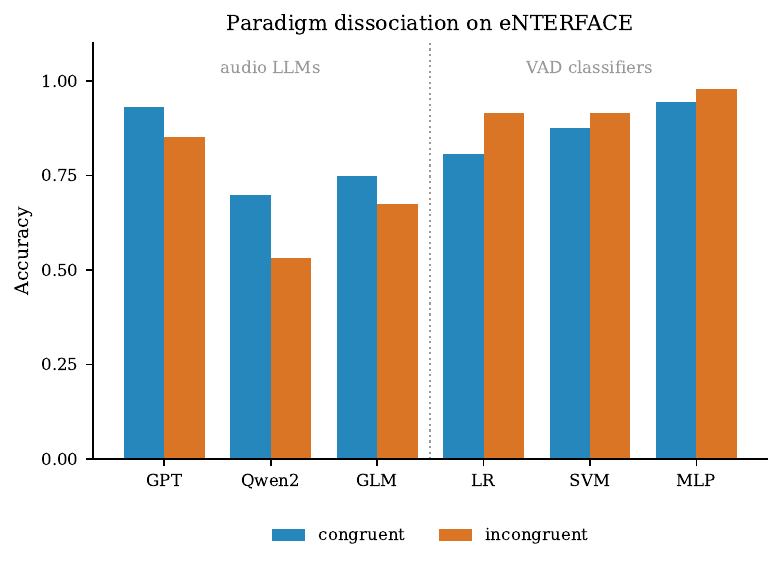}
  \caption{Paradigm dissociation on eNTERFACE (test split, T2 text+audio for the
  LLMs). Audio LLMs lose accuracy on
  incongruent utterances, whereas discriminative classifiers trained on the same
  six VAD features \emph{gain} accuracy, because a larger cross-modal gap makes
  classes more separable. This is consistent with the harm arising during raw
  multimodal integration rather than from the VAD features (the baselines are
  supervised, however, so this is not by itself proof of mechanism).}
  \label{fig:paradigm}
\end{figure}

To locate the effect, we trained discriminative baselines on the same six VAD
features ($\mathbf{x}=[v_t,a_t,d_t,v_a,a_a,d_a]$): logistic regression (LR), a
linear SVM, and an MLP. These reach high accuracy (LR $0.85$, SVM $0.89$, MLP
$0.96$) and, crucially, are \emph{more} accurate on incongruent than congruent
utterances, because a large cross-modal VAD gap separates classes in feature space
(Fig.~\ref{fig:paradigm}).
LLMs given numeric VAD triplets instead of raw signals perform near chance
($0.30$--$0.32$) and show the same benign-incongruence direction. The
detrimental effect of incongruence is thus specific to end-to-end models that
jointly process raw text and audio, not a property of the VAD signal. This
dissociation is consistent with the hypothesis that the effect arises during raw
multimodal evidence integration rather than from VAD feature geometry alone; we
note, however, that the baselines are supervised and the LLMs prompted, so the
contrast is not by itself proof of an integration mechanism.

\subsection{The penalty requires competent integration: GLM-4-Voice}
\label{sec:glm}
That the effect depends on integration is reinforced by the one model that does
not integrate audio well. On eNTERFACE, GLM-4-Voice is the only system whose \pin{}
coefficients change sign with modality, and a diagnostic explains why. It is not an
output problem (96.6\% of its responses parse). Rather, adding audio fails to
help it: accuracy falls from $0.686$ (text-only) to $0.627$ (text+audio) under
T1 and is flat under T2 ($0.729\to0.720$), in contrast to GPT-audio, which climbs
to $0.907$ with audio. Yet GLM is not ignoring the acoustic channel: supplying
audio changes $34$--$38\%$ of its predictions (text-only vs text+audio agreement
$0.66$ and $0.62$). It therefore perturbs a third of its answers on the basis of
audio without improving them, so the relationship between incongruence and error
becomes noisy and direction-unstable rather than systematically harmful (and its
audio-only configuration collapsed entirely in piloting). The contrast is
informative: GPT-audio and Qwen2-Audio integrate the channels well enough that
their conflict reliably \emph{costs} accuracy, whereas GLM-4-Voice integrates so
poorly that conflict merely adds noise. Cross-modal incongruence is thus a
failure mode of \emph{competent} multimodal integration: a model must actually
use both channels for their disagreement to hurt it.

\subsection{MELD: a modest penalty in conversational speech}
\label{sec:meld}
On conversational MELD (GPT-audio, seven classes) the effects are smaller than on
eNTERFACE but point the same way (Table~\ref{tab:meld}). With audio, high-\pin{}
utterances are misclassified more often under both prompts (logistic $p=.032$ and
$p=.017$). Without audio the picture depends on the prompt: the minimal prompt shows
no association ($p=.23$) while the rule-guided one does ($p=.022$). The added-audio
interaction is positive but not significant ($p=.45$ and $p=.87$), so MELD does not
show a clean separation between modalities; what it shows is a small penalty that is
most consistent when the model hears the speaker. Within classes, neutral utterances
are harmed (OR $1.50$) and anger is helped (OR $0.66$), in line with how distinctive
angry prosody is.

\begin{table}[t]
\caption{MELD test split using GPT-audio in all four conditions. OR is the
high- versus low-\pin{} error odds ratio and $\beta$ is the condition-wise
continuous logistic coefficient. The added-audio interaction is not
significant for either prompt. $^{*}p<0.05$.}
\label{tab:meld}
\centering\footnotesize
\resizebox{\ifdim\width>\columnwidth \columnwidth\else \width\fi}{!}{%
\begin{tabular}{lccccccc}
\toprule
\textbf{Cond.} & \textbf{Acc.} & \textbf{F1} & \textbf{Acc$_c$} &
\textbf{Acc$_i$} & \textbf{OR} & $\boldsymbol{\beta}$ & \textbf{logit $p$} \\
\midrule
T1 text-only  & .597 & .479 & .605 & .582 & 1.10 & $+.05$ & .234 \\
T2 text-only  & .606 & .487 & .626 & .569 & 1.27 & $+.10$ & .022$^{*}$ \\
T1 text+audio & .580 & .483 & .597 & .549 & 1.21 & $+.10$ & .032$^{*}$ \\
T2 text+audio & .540 & .469 & .560 & .504 & 1.25 & $+.11$ & .017$^{*}$ \\
\bottomrule
\end{tabular}}
\end{table}

\subsection{MOSEI: the effect holds in spontaneous speech}
\label{sec:mosei}
MOSEI matters because it is the only genuinely spontaneous corpus of the three. On
MOSEI (GPT-audio, six classes) \pin{} predicts error in all four conditions, text-only
and text+audio alike (per-condition $p\le.011$; pooled adjusted OR $1.22$ and $1.20$,
both $q<.001$; Table~\ref{tab:rigor}), and the added-audio interaction is null. The
effect is therefore not confined to acted speech. Within classes, fear, anger and
sadness are helped (OR $0.55$, $0.68$ and $0.50$), while happiness, by far the most
frequent label, is harmed (OR $1.65$).

\subsection{Robustness: controls, repeated measures, and multiple comparisons}
\label{sec:rigor}
This pooled, controlled model is our \emph{primary} inferential analysis; the
per-condition tests reported above are descriptive and should be read through it.
Because the same utterances are scored under multiple conditions and many tests
are reported, we estimate the effect with a single cluster-robust logistic
model per corpus (GPT-audio), pooling the four conditions:
$\text{error}\sim \pin{}\times\text{audio} + \text{prompt} + C(\text{emotion}) +
\text{len} + \text{dur} + \text{tproto} + \text{aproto}$, with \pin{} and all
continuous covariates $z$-standardized (so each OR is per standard deviation),
standard errors clustered by speaker (eNTERFACE, MELD) or utterance (MOSEI), and
Benjamini--Hochberg false-discovery-rate (FDR) correction across the family of
\pin{} effects (Table~\ref{tab:rigor}, Fig.~\ref{fig:forest}; per-cell sample sizes
in Table~\ref{tab:samplesizes}). Here $C(\text{emotion})$ absorbs
class composition and \texttt{tproto}/\texttt{aproto} are the distances to the
nearest text/audio class prototype, a direct control for feature-space geometry
and sample typicality. After these controls the effect sizes are modest
(adjusted OR $\approx1.2$, versus $\approx1.8$ unadjusted) but survive FDR:
\pin{} raises the odds of error in the \emph{text+audio} condition of every corpus
(eNTERFACE OR $1.25$, $q=.022$; MOSEI $1.20$, $q<.001$; MELD $1.15$, $q=.017$). The
text-only picture is heterogeneous: positive on MOSEI (OR $1.22$, $q<.001$), null on
MELD (OR $1.11$, $q=.051$), and \emph{negative} on eNTERFACE (OR $0.85$, $q=.029$),
where a large cross-modal gap is associated with \emph{fewer} text-only errors. With
the model now held constant across modalities, the added-audio interaction is no
longer significant in the conversational corpora (MELD $p=.45/.87$;
MOSEI $p=.73/.62$).
Prototype distance is itself a significant predictor, yet \pin{}
remains significant after adjusting for it: cross-modal incongruence predicts LLM
error \emph{beyond} the feature-space separability that benefits the
discriminative baselines. The LLM penalty is therefore not reducible to general
VAD geometry.

\subsection{Model or modality?}
\label{sec:modelabl}
A text-only condition can be served by any language model, and that choice turns out
to matter. We therefore ran the condition twice, holding the transcripts, the
prompts, the incongruence scores and the labels fixed, and varying only the reader:
\texttt{gpt-5.2}, a text-only model, against \texttt{gpt-audio-1.5}, the
audio-capable model used throughout the main analysis.

The two readers are equally accurate, differing by at most one and a half points on
any corpus, yet they are not equally sensitive to incongruence
(Table~\ref{tab:modelabl}). On MELD and on MOSEI they agree, both flat on the former
and both clearly positive on the latter. On eNTERFACE they part company. The text
model shows nothing at all, while the audio-capable model shows a protective
association strong enough to exclude unity.

The substantive reading is that the coupling between incongruence and error is not a
property of the stimulus alone. Two models of comparable accuracy, given identical
text, can lean in opposite directions, which is what an integration account would
predict and a purely lexical one would not. The methodological reading is a caution.
Pairing a text-only condition with a different model from the multimodal condition
confounds model with modality, and on eNTERFACE that confound is large enough to
reverse the sign of the estimate. The main analysis therefore holds the model fixed.
Two limits temper the inference: \texttt{gpt-5.2} was never run with audio, so the
design is not fully crossed, and the two models differ in more than audio capability.

\begin{table}[t]
\caption{Text-only condition read by two different models, with transcripts, prompts,
\pin{} and labels held fixed. Accuracy is materially the same; sensitivity to
incongruence is not. Odds ratios come from a model fitted to the text-only condition
alone, so they differ slightly from the text-only cells of Table~\ref{tab:rigor},
which come from the pooled model with an audio interaction.}
\label{tab:modelabl}
\centering\footnotesize
\resizebox{\ifdim\width>\columnwidth \columnwidth\else \width\fi}{!}{%
\begin{tabular}{llcc}
\toprule
\textbf{Corpus} & \textbf{Reader} & \textbf{Acc.} & \textbf{Adj.\ OR (95\% CI)} \\
\midrule
\multirow{2}{*}{eNTERFACE} & \texttt{gpt-5.2}       & $.832$ & $1.12$ $[0.91,1.37]$ \\
                           & \texttt{gpt-audio-1.5} & $.818$ & $\mathbf{0.78}$ $[0.63,0.97]$ \\
\midrule
\multirow{2}{*}{MELD}      & \texttt{gpt-5.2}       & $.594$ & $1.03$ $[0.94,1.13]$ \\
                           & \texttt{gpt-audio-1.5} & $.601$ & $1.06$ $[0.94,1.18]$ \\
\midrule
\multirow{2}{*}{MOSEI}     & \texttt{gpt-5.2}       & $.543$ & $\mathbf{1.23}$ $[1.14,1.33]$ \\
                           & \texttt{gpt-audio-1.5} & $.536$ & $\mathbf{1.24}$ $[1.14,1.35]$ \\
\bottomrule
\end{tabular}}
\end{table}

\subsection{Cross-model generality on the extension corpora}
\label{sec:crossmodel}
\begin{figure}[t]
  \centering
  \includegraphics[width=\columnwidth]{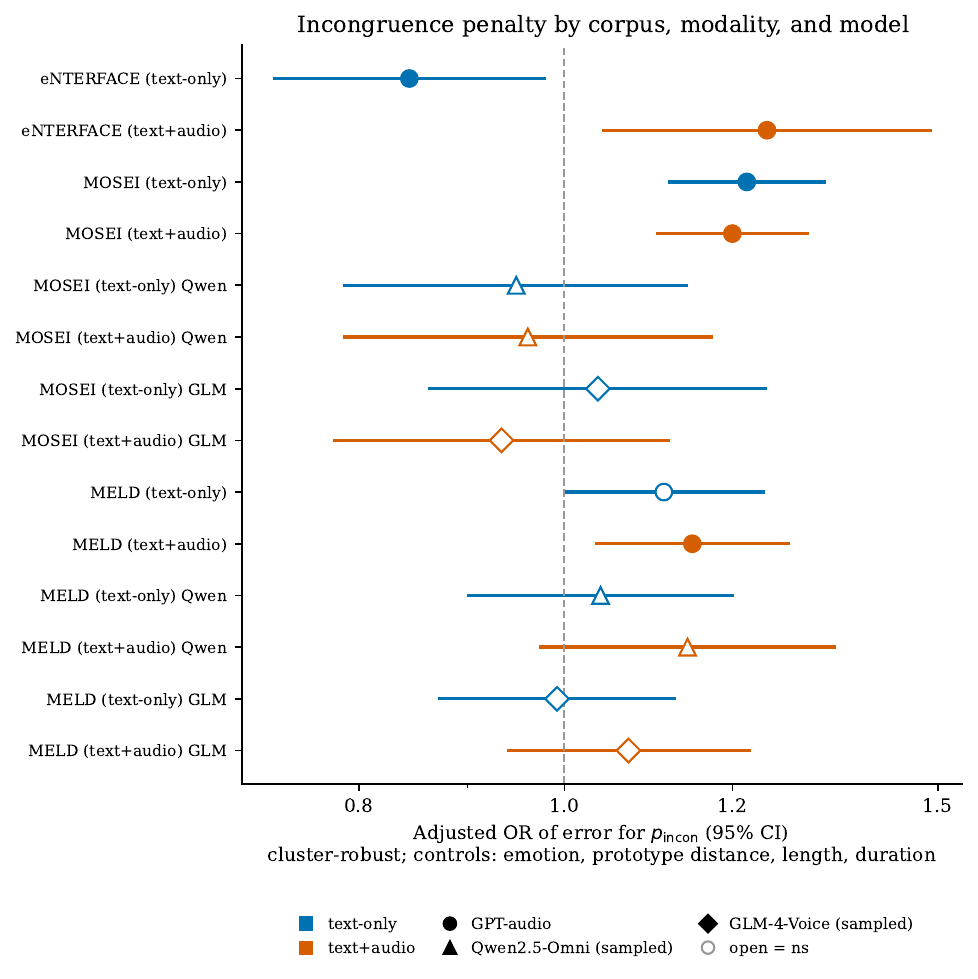}
  \caption{Adjusted association between \pin{} and error from the cluster-robust
  logistic model, with controls for emotion, prototype distance, length and duration
  (95\% CIs). Circles are GPT-audio on all three corpora; triangles are Qwen2.5-Omni
  and diamonds GLM-4-Voice on the matched MELD and MOSEI samples. Filled markers have
  intervals that exclude $1$. For GPT-audio the text+audio association is positive in
  every corpus, while the text-only association changes sign across corpora.
  Qwen2.5-Omni agrees in direction on MELD ($1.14$) without reaching significance, and
  GLM-4-Voice shows no significant association, in line with its much weaker
  accuracy.}
  \label{fig:forest}
\end{figure}

We evaluate all three audio-LLM families on MELD and MOSEI under the identical
cluster-robust specification (Table~\ref{tab:crossmodel}, Fig.~\ref{fig:forest}),
with GPT-audio on the full corpora and Qwen2.5-Omni and GLM-4-Voice on a matched
stratified sample (provider rate limits). Three patterns emerge. First, the
\textbf{direction is shared but the significance is carried by GPT-audio}: in the
text+audio condition all three families give point estimates above~$1$ on MELD
($1.15$, $1.14$, $1.07$), but only GPT-audio's interval excludes~$1$. We therefore
state the cross-corpus text+audio effect as \emph{consistent in direction across
model families and statistically established for GPT-audio}, rather than as a
replicated significant effect. Second, \textbf{the MOSEI effect is GPT-specific}: it
is strong for GPT in both modalities (OR $1.22$/$1.20$, $q<.001$) but absent for Qwen
and GLM (CIs spanning $1$), so we do not over-interpret it. Third, \textbf{model
competence bounds the effect}: GLM-4-Voice is a much weaker classifier (accuracy $28$--$37\%$ on MELD and
$30$--$34\%$ on MOSEI, versus $51$--$62\%$ for Qwen and GPT, and audio even
\emph{lowers} its MELD accuracy), and its incongruence effect is non-significant on
both corpora. Consistent with its inconsistency on eNTERFACE, GLM-4-Voice is best
read as too weak at the underlying SER task to resolve the effect rather than as
counter-evidence. The Qwen and GLM runs use a stratified sample and thus have lower
power than the full-corpus GPT analysis.

\begin{table}[t]
\caption{Cross-model incongruence effect on the extension corpora: adjusted OR of
error for \pin{} (95\% CI) from the cluster-robust model, for all three audio-LLM
families. GPT-audio is on the full corpora; Qwen2.5-Omni and GLM-4-Voice on the
matched stratified sample. Bold = CI excludes $1$.}
\label{tab:crossmodel}
\centering\footnotesize
\setlength{\tabcolsep}{4pt}
\resizebox{\ifdim\width>\columnwidth \columnwidth\else \width\fi}{!}{%
\begin{tabular}{llccc}
\toprule
\textbf{Corpus} & \textbf{Modality} & \textbf{GPT-audio} & \textbf{Qwen2.5-Omni} & \textbf{GLM-4-Voice} \\
\midrule
\multirow{2}{*}{MELD}  & text-only  & $1.11$ $[1.00,1.24]$ & $1.04$ $[0.90,1.20]$ & $0.99$ $[0.87,1.13]$ \\
                       & text+audio & $\mathbf{1.15}$ $\mathbf{[1.04,1.28]}$ & $1.14$ $[0.97,1.34]$ & $1.07$ $[0.94,1.22]$ \\
\midrule
\multirow{2}{*}{MOSEI} & text-only  & $\mathbf{1.22}$ $\mathbf{[1.12,1.33]}$ & $0.95$ $[0.79,1.14]$ & $1.04$ $[0.86,1.24]$ \\
                       & text+audio & $\mathbf{1.20}$ $\mathbf{[1.11,1.30]}$ & $0.96$ $[0.79,1.17]$ & $0.93$ $[0.78,1.12]$ \\
\bottomrule
\end{tabular}}
\end{table}

\begin{table}[t]
\caption{Adjusted effect of \pin{} on GPT-audio error from a cluster-robust
logistic model per corpus (controls: prompt, ground-truth emotion, text/audio
prototype distance, text length, audio duration; SEs clustered by speaker or
utterance). Odds ratios with 95\% CIs; $q$ is the Benjamini--Hochberg
FDR-corrected $p$-value across the six effects. The text+audio effect (bold) is
harmful and significant in all three corpora; the text-only effects are inconsistent
in sign and are model-dependent (Section~\ref{sec:modelabl}).}
\label{tab:rigor}
\centering\footnotesize
\resizebox{\ifdim\width>\columnwidth \columnwidth\else \width\fi}{!}{%
\begin{tabular}{llccc}
\toprule
\textbf{Corpus} & \textbf{Modality} & \textbf{Adj.\ OR} & \textbf{95\% CI} & \textbf{$q$ (FDR)} \\
\midrule
\multirow{2}{*}{eNTERFACE} & text-only  & $0.85$ & $[0.73,0.98]$ & $.029$ \\
                           & text+audio & $\mathbf{1.25}$ & $[1.04,1.49]$ & $.022$ \\
\midrule
\multirow{2}{*}{MOSEI}     & text-only  & $1.22$ & $[1.12,1.33]$ & $<.001$ \\
                           & text+audio & $\mathbf{1.20}$ & $[1.11,1.30]$ & $<.001$ \\
\midrule
\multirow{2}{*}{MELD}      & text-only  & $1.11$ & $[1.00,1.24]$ & $.051$ (ns) \\
                           & text+audio & $\mathbf{1.15}$ & $[1.04,1.28]$ & $.017$ \\
\bottomrule
\end{tabular}}
\end{table}

\subsection{Does lexical transparency govern the effect?}
\label{sec:crosscorpus}
The three corpora differ sharply in how much the transcript alone reveals the
emotion. eNTERFACE is the most explicit of them by every measure we have, the
lexicon-based one and the two VAD-independent ones alike, while MELD and MOSEI sit
close to their floors (Table~\ref{tab:crosscorpus}). A tempting account of the
cross-corpus pattern is that this lexical transparency decides where incongruence
hurts.

The data do not bear it out. The text+audio penalty holds in all three corpora at
much the same magnitude (OR $1.15$ to $1.25$), from the most transparent to the
least, and the text-only coefficients do not follow transparency either. eNTERFACE,
the most transparent corpus, is protective; of the other two, MOSEI is positive and
MELD null, although the measures disagree on which of the two is the more
transparent.

Transparency remains a useful description of these corpora. It is why audio rescues
a misleading transcript in almost a quarter of incongruent MELD utterances but in
only about three per cent of eNTERFACE ones. It is not, on this evidence, an
explanation of where the incongruence penalty falls. The differences that remain are
better attributed to context: MELD is multi-party dialogue whose emotion often turns
on conversational history that an utterance in isolation does not carry.

\begin{table*}[t]
\caption{Lexical emotional transparency by three measures: a lexicon-based one
(text-VAD nearest-prototype accuracy, which shares signal with \pin{}) and two
VAD-\emph{independent} ones (transcript-only GPT accuracy, and the cross-validated
macro-F1 of a linear probe on frozen all-mpnet-base-v2 sentence embeddings). All
three agree that eNTERFACE is by far the most explicit. The text-only effect (last
column) does not order with transparency, which is why we do not treat transparency
as governing it. The
embedding probe reaches a ceiling on eNTERFACE, whose closed scripted-sentence set
makes the transcript trivially diagnostic.}
\label{tab:crosscorpus}
\centering\footnotesize
\resizebox{\linewidth}{!}{%
\begin{tabular}{lccccc}
\toprule
\textbf{Corpus} & \textbf{Register} & \textbf{Text$\to$label (VAD)} &
\textbf{Transcript-only GPT acc} & \textbf{Embed.\ probe F1} & \textbf{Text-only effect} \\
\midrule
eNTERFACE'05 & read, emotion-prototypical & $0.733$ & $0.818$ & $1.00$ & protective \\
CMU-MOSEI    & spontaneous monologue       & $0.388$ & $0.536$ & $0.24$ & present \\
MELD         & conversational dialogue     & $0.188$ & $0.601$ & $0.27$ & absent \\
\bottomrule
\end{tabular}}
\end{table*}

\subsection{Robustness checks}
\label{sec:robust}
We subject the score and the effect to several stress tests. \textbf{Score
construction.} The ranking of \pin{} is essentially invariant to the GMM
configuration: refitting with two or three components, full/diagonal/tied
covariance, and several seeds yields scores that correlate with the reported one
at Spearman $\rho=1.00$. The congruent/incongruent contrast is also stable across
split points from $0.3$ to $0.7$ (peaking near $0.5$). Re-deriving \pin{} from
cosine or Mahalanobis distances rather than Euclidean produces correlated scores
(Spearman $0.59$ and $0.81$); how these and other simple scores compare as
stratifiers is examined below. \textbf{Mixed-effects estimation.} A
Bayesian generalised linear mixed model with a random intercept per speaker
(eNTERFACE, pooled GPT) gives \pin{} posterior OR $=1.11$, 95\% credible interval
$[1.03,1.20]$; the pooled estimate is attenuated relative to the text+audio cell
because it averages over the text-only condition, whose eNTERFACE coefficient is
protective. Because each utterance is scored under four
conditions, we also fit a crossed-random-effects model with intercepts for both
utterance and speaker ($n=5105$ observations, $1287$ utterances, $43$ speakers);
the modality pattern sharpens rather than disappears: the text-only slope is null
(posterior OR $=1.10$, $[0.95,1.26]$) while the \pin{}$\times$audio interaction is
clearly positive (posterior log-OR $=+0.48$, SD $0.11$), implying a text+audio slope
of OR $\approx1.78$. The text+audio penalty is therefore not an artefact of the
repeated-measures structure.

\textbf{A second audio-VAD extractor.} Because \pin{} depends on the audio-VAD
model, we recompute it with an independent extractor (a WavLM model trained for
the MSP Odyssey dimensional task) whose outputs correlate with the primary
extractor (per-dimension $|r|=0.77$--$0.90$) but which shares no architecture or
training data. The resulting score correlates $0.39$--$0.58$ with the primary
\pin{}, and agreement between the two depends on the corpus
(Table~\ref{tab:secondext}). On MOSEI the two extractors agree closely and every
estimate excludes $1$. On MELD neither yields a significant association under this
reduced specification, which controls for emotion class but not for prototype
distance, length or duration. On eNTERFACE both reproduce the protective text-only association, but neither
reproduces the text+audio penalty without the full control set. We therefore treat
the score as robust to the choice of extractor on MOSEI and only partly so elsewhere
(Section~\ref{sec:limits}).

\textbf{Comparison with simpler stratifiers.} An evaluator who wants to separate
congruent from incongruent utterances has cheaper options than a fitted mixture, so we
compared \pin{} with six of them (Table~\ref{tab:stratifiers}): the raw Euclidean
distance, the distance after z-scoring each modality, cosine distance, the absolute
valence gap, a flag for text and audio valence of opposite sign, and disagreement
between the nearest class prototype of each modality. (Disagreement between
off-the-shelf text and audio emotion classifiers was also tried on eNTERFACE, but the
audio classifier performed below chance there, so we do not report it.) Each score was standardised and
entered in place of \pin{} in the primary model, so the odds ratios share a scale.
Three things stand out. First, no score dominates. Cosine distance, the valence flip
and prototype disagreement predict text+audio error more strongly than \pin{} on
eNTERFACE and MOSEI, but cosine distance and the valence flip find nothing on MELD,
where \pin{} and the raw distance are the strongest of all. \pin{} is the only score
that needs no labels and is significant in all three corpora. Second, strength is not
specificity. The valence flip predicts eNTERFACE text-only error as strongly as
text+audio error (OR $1.82$ and $1.70$), so it partly marks transcripts that mislead
on their own rather than a conflict between the channels. Prototype disagreement is
both strong and specific to audio on eNTERFACE, but it needs labelled class
prototypes, which rules it out for unlabelled data and for reliability gating at
deployment. Third, the mixture contributes a threshold rather than a ranking: \pin{}
orders utterances exactly as the raw distance does, and the posterior transform
strengthens the eNTERFACE estimate ($1.25$ against $1.10$) while slightly weakening
the MOSEI one ($1.20$ against $1.23$). We therefore offer \pin{} as a label-free
default, with prototype disagreement as a complementary check where labels exist, and
do not claim that it is the strongest predictor of error.

\begin{table}[t]
\caption{\pin{} against simpler stratifiers. Each score replaces \pin{} in the primary
model of Table~\ref{tab:rigor}; entries are GPT-audio text+audio odds ratios per
standard deviation (binary flags included). $^{*}$95\% CI excludes $1$. ``Labels'' marks
scores that require gold emotion labels; prototype disagreement is cross-fitted across
speakers, so no utterance's flag uses its own speaker's labels.}
\label{tab:stratifiers}
\centering\footnotesize
\resizebox{\ifdim\width>\columnwidth \columnwidth\else \width\fi}{!}{%
\begin{tabular}{lcccc}
\toprule
\textbf{Score} & \textbf{Labels} & \textbf{eNTERFACE} & \textbf{MELD} & \textbf{MOSEI} \\
\midrule
\pin{} (GMM posterior) & no & $\mathbf{1.25}^{*}$ & $\mathbf{1.15}^{*}$ & $\mathbf{1.20}^{*}$ \\
Euclidean distance      & no & $1.10$       & $1.15^{*}$ & $1.23^{*}$ \\
z-scored distance       & no & $1.37^{*}$   & $1.00$     & $1.21^{*}$ \\
Cosine distance         & no & $1.52^{*}$   & $1.00$     & $1.32^{*}$ \\
Valence gap             & no & $1.11$       & $0.96$     & $1.18^{*}$ \\
Valence-sign flip       & no & $1.70^{*}$   & $0.91$     & $1.31^{*}$ \\
Prototype disagreement  & yes & $2.16^{*}$  & $1.07^{*}$ & $1.31^{*}$ \\
\bottomrule
\end{tabular}}
\end{table}

\begin{figure}[t]
  \centering
  \includegraphics[width=\linewidth]{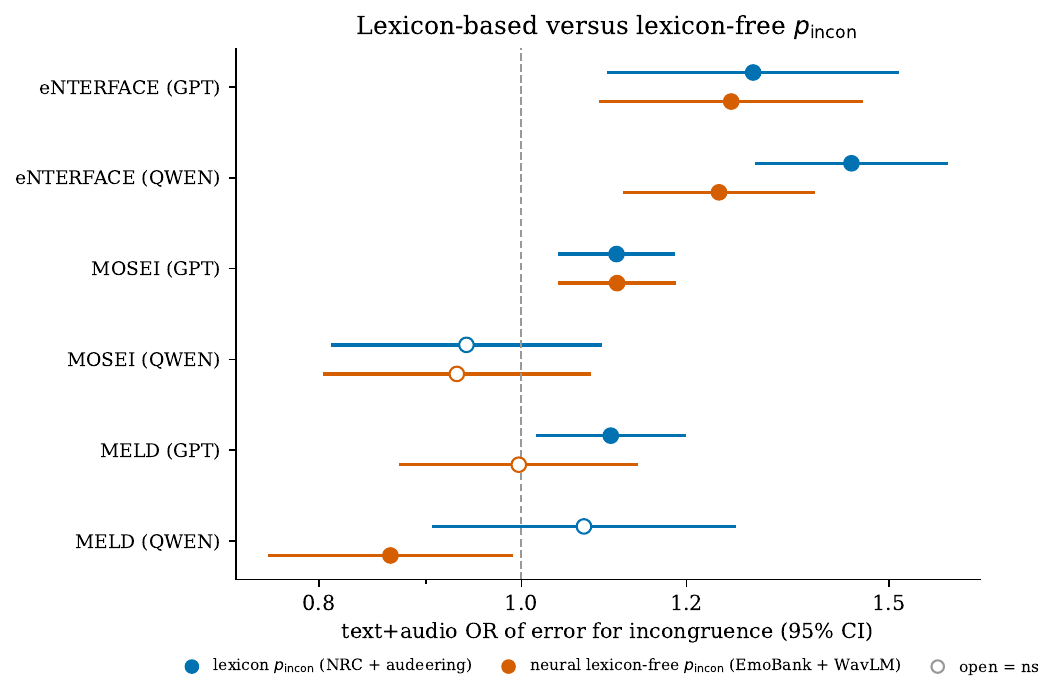}
  \caption{Convergent validity: the text+audio incongruence effect under the
  lexicon-based \pin{} (NRC lexicon + audeering audio-VAD; blue) versus a fully
  neural, lexicon-free \pin{} (EmoBank-trained text-VAD regressor + WavLM audio-VAD;
  orange), under the identical cluster-robust specification. Filled markers exclude
  OR${=}1$. On eNTERFACE and MOSEI the neural score reproduces the text+audio
  association found with the lexicon-based one; on MELD it does not, so the
  convergence is partial.}
  \label{fig:lexfree}
\end{figure}

\textbf{A fully lexicon-free reconstruction.} We replace both VAD channels
with independent neural models: an EmoBank-supervised text-VAD regressor and
the WavLM audio-VAD extractor (Fig.~\ref{fig:lexfree}). The resulting item-level score correlates only
weakly with the lexicon-based \pin{} (Spearman $0.05$--$0.20$), so the two
operationalisations are not interchangeable. Nevertheless, the positive
text+audio association reappears on eNTERFACE for GPT-audio
(OR $1.26$ $[1.09,1.46]$) and Qwen
(OR $1.24$ $[1.12,1.38]$), and on MOSEI for GPT-audio
(OR $1.11$ $[1.04,1.18]$). It does not reappear for GPT-audio on MELD;
Qwen on MELD instead shows a protective association
(OR $0.87$ $[0.76,0.99]$). These results provide partial convergent validity,
while showing that the estimated relationship can depend on the
operationalisation, corpus, and model family.

\textbf{Conversational context.} MELD labels were assigned with the dialogue in
view, yet we classify each utterance on its own. To see how much this matters we
re-ran GPT-audio on a class-balanced sample of $704$ MELD utterances with the
preceding turns supplied. Context raises accuracy by three to four points (text-only
$0.548\!\to\!0.575$; text+audio $0.523\!\to\!0.565$). In this balanced sample,
however, \pin{} is unrelated to error with or without context (all ORs between $0.93$
and $0.96$, $p>.3$), so it cannot tell us whether context would remove the
full-corpus association. A likely reason is that balancing the classes down-weights
neutral utterances, which carry most of the MELD effect
(Table~\ref{tab:peremotion}).

\textbf{Unparseable responses.} Invalid (non-label) outputs are rare on eNTERFACE
($\le1.5\%$) and on MELD text-only ($0\%$), but reach $5\%$ on MELD text+audio and
$\sim15\%$ on MOSEI in \emph{both} modalities (Table~\ref{tab:samplesizes}); our
primary analysis excludes them. As a sensitivity check we instead count every invalid
answer as an error and refit. MELD is unaffected in either modality (text-only
$1.07\!\to\!1.08$; text+audio $1.10\!\to\!1.10$). Where refusals are common the
text+audio estimate attenuates (eNTERFACE $1.29\!\to\!1.11$ $[1.00,1.23]$; MOSEI
$1.11\!\to\!1.02$ $[0.97,1.08]$, which loses significance), so part of the
MOSEI signal reflects the model declining to answer on hard, high-\pin{} items. The
MELD text+audio effect is thus the most robust to how refusals are treated, and the
MOSEI one the least.

\textbf{MOSEI label construction.} Reducing MOSEI's multi-label emotion intensities
to a single argmax label is a simplification, so we test its influence by recovering
the original intensity vectors (matched to every utterance by transcript; the argmax
reproduces the released label for $99.2\%$ of utterances). Restricting to
unambiguously labelled utterances does not weaken the effect; if anything it
strengthens it: the text+audio association grows from OR $1.11$ on all utterances
to $1.13$ on those with a unique top emotion and $1.28$ on high-intensity,
unambiguous ones ($12.7\%$ of utterances have a tied top emotion).
The argmax reduction is therefore conservative rather than a source of spurious
signal.

\textbf{Utterance selection in MELD.} Lexical VAD is undefined for $536/2610$
($20.5\%$) of MELD test utterances, which are consequently excluded. These are
overwhelmingly very short turns (mean $2.4$ words vs.\ $9.7$ for included
utterances), mostly backchannels and interjections, and their gold labels skew toward
surprise ($19\%$ vs.\ $9\%$) and away from anger, sadness and disgust. The MELD
results therefore describe lexically contentful turns and should not be read as
representative of single-word backchannels; we note this as a scope restriction.

\begin{table}[t]
\caption{Association between \pin{} and GPT-audio error under the primary
audio-VAD extractor and an independent WavLM extractor. Entries are adjusted
ORs with 95\% CIs from the same reduced cluster-robust specification.
Bold entries have CIs excluding $1$. Robustness is strongest on MOSEI and
limited on MELD and eNTERFACE.}
\label{tab:secondext}
\centering\footnotesize
\resizebox{\ifdim\width>\columnwidth \columnwidth\else \width\fi}{!}{%
\begin{tabular}{llcc}
\toprule
\textbf{Corpus} & \textbf{Modality} & \textbf{Primary} & \textbf{WavLM} \\
\midrule
\multirow{2}{*}{eNTERFACE}
 & text-only  & $\mathbf{0.715}$ $\mathbf{[0.598,0.853]}$ & $\mathbf{0.827}$ $\mathbf{[0.708,0.966]}$ \\
 & text+audio & $1.006$ $[0.829,1.221]$ & $1.140$ $[0.970,1.340]$ \\
\midrule
\multirow{2}{*}{MELD}
 & text-only  & $1.046$ $[0.955,1.146]$ & $1.008$ $[0.928,1.094]$ \\
 & text+audio & $1.077$ $[0.989,1.173]$ & $0.889$ $[0.786,1.005]$ \\
\midrule
\multirow{2}{*}{MOSEI}
 & text-only  & $\mathbf{1.104}$ $\mathbf{[1.025,1.188]}$ & $\mathbf{1.140}$ $\mathbf{[1.059,1.227]}$ \\
 & text+audio & $\mathbf{1.085}$ $\mathbf{[1.011,1.164]}$ & $\mathbf{1.084}$ $\mathbf{[1.010,1.163]}$ \\
\bottomrule
\end{tabular}}
\end{table}

\begin{table}[t]
\caption{GPT-audio analysis sample sizes (pooled over the two prompts) and
invalid-response rates. Invalid (non-label) outputs are negligible on eNTERFACE and
MELD but frequent on MOSEI in \emph{both} modalities, the audio-capable model
declining to answer on roughly one seventh of those utterances; the invalid-as-error
sensitivity check (Section~\ref{sec:robust}) addresses them.}
\label{tab:samplesizes}
\centering\footnotesize
\resizebox{\ifdim\width>\columnwidth \columnwidth\else \width\fi}{!}{%
\begin{tabular}{llccc}
\toprule
\textbf{Corpus} & \textbf{Modality} & \textbf{$N$ scored} & \textbf{$N$ valid} & \textbf{invalid \%} \\
\midrule
\multirow{2}{*}{eNTERFACE} & text-only  & $2574$ & $2568$ & $0.2$ \\
                           & text+audio & $2574$ & $2537$ & $1.4$ \\
\midrule
\multirow{2}{*}{MELD}      & text-only  & $4148$ & $4147$ & $0.02$ \\
                           & text+audio & $4148$ & $3931$ & $5.2$ \\
\midrule
\multirow{2}{*}{MOSEI}     & text-only  & $8422$ & $7185$ & $14.7$ \\
                           & text+audio & $8422$ & $7195$ & $14.6$ \\
\bottomrule
\end{tabular}}
\end{table}

\begin{table}[t]
\caption{Illustrative eNTERFACE utterances, with the nearest-prototype emotion of
each modality, \pin{}, and the GPT-audio prediction (T2, text+audio). Category and
score capture different things: the first row has discordant prototype labels but a
low score, the third concordant labels but a high one, because \pin{} measures
distance in VAD space rather than a change of category.}
\label{tab:examples}
\centering\footnotesize
\resizebox{\ifdim\width>\columnwidth \columnwidth\else \width\fi}{!}{%
\begin{tabular}{lllcl}
\toprule
\textbf{Transcript} & \textbf{Gold} & \textbf{text/audio} & \pin{} & \textbf{GPT} \\
\midrule
``You're getting paid to work, not drink coffee!'' & anger & anger/sadness & $0.20$ & anger \\
``Aaaaah a cockroach!!!'' & disgust & disgust/sadness & $0.76$ & fear \\
``Everything was so perfect! I just don't understand!'' & sadness & sadness/sadness & $0.81$ & sadness \\
``Oh no, tell me this is not true, please!'' & sadness & anger/sadness & $1.00$ & fear \\
\bottomrule
\end{tabular}}
\end{table}

\section{User Study: Validating \pin{}}
\label{sec:user}
To validate the automatic score against human perception, two English-speaking
raters judged 180 eNTERFACE utterances, sampled by a stratified design (six
emotions $\times$ three \pin{} buckets $\times$ ten, at most six per speaker), as
congruent or incongruent given both transcript and audio. Inter-rater
reliability was moderate (Cohen's $\kappa=0.46$, Krippendorff's $\alpha=0.46$),
reflecting the genuine subjectivity of incongruence perception. Agreement between
each rater and the automatic metric was substantially higher (Krippendorff's
$\alpha=0.70$ and $0.73$; three-way $\alpha=0.63$). In the low-\pin{} bucket
raters labelled $15$--$17\%$ of utterances incongruent, versus $85$--$92\%$ in
the mid and high buckets, and \pin{} correlates monotonically with human
judgement (Spearman $r_s=0.65$, $p<0.001$). The moderate inter-rater agreement
is expected for this construct and motivates a continuous, probabilistic score
rather than a hard binary label.

We extend the same protocol to the conversational and spontaneous corpora,
collecting congruent/incongruent judgements on stratified MELD and CMU-MOSEI samples
of $120$ utterances each. Agreement falls in the same moderate range as
eNTERFACE: inter-rater reliability of Krippendorff's $\alpha=0.42$ on MELD and
$0.45$ on MOSEI, with higher metric--human agreement ($\alpha=0.63$ and
$0.67$), and \pin{} again correlates monotonically with human judgement
(Spearman $r_s=0.58$ and $0.62$). Agreement is slightly lower than on
acted speech, as expected for material that is more context-dependent. Across the
three corpora the score therefore tracks perceived incongruence, most strongly where
the emotion is acted and explicit. The computational checks reported above, an independent audio-VAD extractor
and a fully lexicon-free reconstruction, point the same way on eNTERFACE and
MOSEI.

\section{Discussion}
\label{sec:discuss}
Three findings cohere. First, \pin{} predicts LLM error beyond what feature
geometry explains: the association survives cluster-robust estimation, FDR correction
and controls for the ground-truth emotion and for distance to the class prototypes.
Second, the cost appears when the model hears the speaker. With the model held
constant, the text+audio association is positive in all three corpora, whereas the
text-only association changes sign from corpus to corpus and, on eNTERFACE, with the
model that reads the transcript. Third, the features that hurt the LLMs help
supervised classifiers (logistic-regression OR for \pin{} of $0.80$ on text VAD and
$0.82$ on combined VAD), which is consistent with an integration account. We hold
that interpretation loosely: the baselines are supervised and the LLMs prompted, so
the contrast may partly reflect how they were trained, and our data do not expose
the models' internal mechanisms. For practitioners building voice agents the
practical point is simple. Disagreement between what users say and how they say it
is a measurable source of error, so incongruence-stratified evaluation is worth
adding to multimodal SER benchmarks, and a high \pin{} is a reasonable trigger for
reliability gating.

\paragraph{Three constructs, kept distinct}
Our terminology deliberately separates three things that the text-only results
could otherwise conflate. (i) \emph{Stimulus incongruence}: a property of the
utterance itself, namely that its acoustic and verbal emotional signals point in
different directions, which \pin{} operationalises from the two VAD channels. (ii)
\emph{Transcript atypicality/ambiguity}: a property of the words alone, i.e.\ how
emotionally (un)usual or (un)informative the transcript is; this is what a
text-only model can actually perceive. (iii) \emph{Multimodal integration failure}:
a property of a model that receives both channels and must reconcile them. Only
(iii) is ``cross-modal'' in the experienced sense, and we reserve that reading for
the text+audio conditions. A text-only model never hears the audio, so when \pin{}
predicts its errors this cannot be a cross-modal penalty as \emph{experienced by
the model}. A high \pin{} can co-occur with an unusual transcript (ii), and whether
a text-only model is helped or hurt by that depends on the model, as the ablation in
Section~\ref{sec:modelabl} shows. We therefore do not call the text-only association
a ``cross-modal incongruence penalty''. The genuinely cross-modal effect is the
text+audio one.

\paragraph{Where the words say little}
Conversational speech shows why the audio should matter. In a line such as ``Oh,
that's so Monica'' the words are emotionally flat whether the turn is delivered
warmly or with sarcasm, and the distinction lives entirely in the prosody. Such
``audio-rescue'' items, where the voice carries the correct emotion and the
transcript misleads, make up $23.9\%$ of incongruent MELD utterances but only about
$3\%$ of eNTERFACE ones, and they are invisible to any model that reads the
transcript alone. That GPT-audio is still penalised by incongruence when it hears
such utterances suggests that it does not reliably defer to the voice when the two
channels conflict.

\paragraph{Relation to concurrent findings on lexical dominance}
Our account converges with, and sharpens, a recent line of evidence that
audio-LLMs underuse acoustic information. \citet{chen2026listen} report a
consistent ``lexical dominance'' in speech emotion recognition, with models
defaulting to the words and failing under deliberately constructed cue conflict;
\citet{billa2026modality} show that audio-LLMs follow the transcript over a
contradicting voice even when instructed to trust the voice; and acoustic
consistency suites penalise tone-text mismatch \citep{maimon2025salmon}. These
studies establish \emph{that} the failure exists, largely through adversarial or
synthetically conflicted stimuli. Our contribution is complementary and, we
argue, more diagnostic in two respects. First, we do not engineer the conflict:
\pin{} measures \emph{naturally occurring} incongruence on a continuous,
human-validated scale and shows it predicts error in ordinary corpora, so the
effect is not an artefact of adversarial construction. Second, our results suggest
that the problem is not simply that these models ignore the audio. If they did,
incongruence would cost them no more with the voice than without it. Instead, the
cost is consistent only when the voice is supplied, which points to a failure in
reconciling the two channels rather than a failure to perceive one of them.

\paragraph{A note on discrete versus continuous incongruence}
A discrete definition (disagreement between the nearest-prototype emotion of
each modality) agrees only weakly with \pin{} and with human judgement
(Table~\ref{tab:examples} shows how the two come apart on individual utterances), yet
where labelled prototypes are available it predicts error strongly
(Table~\ref{tab:stratifiers}). The two capture different things: the discrete flag
marks utterances whose channels point to different categories, \pin{} how far apart
the channels lie. We see them as complementary rather than competing.

\section{Limitations}
\label{sec:limits}
\textbf{Construct validity of \pin{}.} The human validation uses two raters per
corpus and yields moderate inter-rater agreement throughout, and the eNTERFACE study
is considerably larger than those on MELD and CMU-MOSEI. We therefore do not claim
that \pin{} is a turnkey measure of perceived incongruence across registers: its
perceptual grounding is strongest for read, emotionally explicit speech, and the
conversational and spontaneous corpora would benefit from more raters and from higher
agreement than we obtain here.

\textbf{Choice of score.} \pin{} is not the strongest predictor of error in every
corpus (Table~\ref{tab:stratifiers}); its case rests on needing no labels and on
working in all three corpora. A score that combines it with label-based prototype
disagreement may stratify more sharply, and we leave that to future work.

\textbf{Effect sizes and statistics.} After controls and FDR correction the
effects are modest (adjusted OR $\approx1.1$--$1.2$), as expected for hard,
multi-class tasks; we report them with confidence intervals rather than emphasise
significance alone. Our crossed mixed model includes random intercepts for
utterance and speaker but not for model or prompt; a model crossed over all four
would be preferable once comparable full-corpus data exist for every model.

\textbf{Model and corpus scope.} The full three-model comparison is on eNTERFACE;
on the cross-corpus extension all three families are run on a stratified MELD/MOSEI
sample (GPT-audio also on the full corpora): Qwen2.5-Omni agrees in direction on
MELD without reaching significance and shows no MOSEI effect, while GLM-4-Voice is
too weak at the underlying task to resolve either. A complete three-model run on the \emph{full}
(unsampled) MELD/MOSEI was precluded by provider rate limits and is left to future
work. Lexical VAD is
undefined for roughly one fifth of conversational utterances (short
function-word backchannels); these excluded turns are very short ($2.4$ vs.\ $9.7$
words) and skew toward surprise and away from anger/sadness, so the MELD results
describe lexically contentful turns (Section~\ref{sec:robust}).

\textbf{MELD context dependence.} MELD emotion labels are assigned in conversational
context, but the model classifies utterances in isolation. Supplying the preceding
turns improves accuracy, but our class-balanced context sample was too small, and
too light on neutral utterances, to tell whether context also removes the
incongruence association (Section~\ref{sec:robust}).

\textbf{MOSEI labels.} Reducing MOSEI's multi-label intensities to a single argmax
label is a simplification, but a sensitivity analysis on the recovered intensities
shows the effect is, if anything, stronger on unambiguously labelled utterances
(Section~\ref{sec:robust}), so it does not arise from argmax noise.

\textbf{Size of the effect.} The robustness checks (Section~\ref{sec:robust}) show
that the score and the paradigm dissociation are stable, but that the size of the
effect depends on corpus, extractor and model: the MELD effect does not survive a
change of audio-VAD extractor, the MOSEI effect does not replicate with a second
model family, and the eNTERFACE text+audio penalty needs the full control set to
reach significance. We therefore present the effect as consistent in direction
rather than uniform in size. Finally, although we validate \pin{} with a
second audio-VAD extractor, both extractors are trained on dimensional MSP data
and may share corpus-specific biases.

\section{Conclusion}
\label{sec:conclusion}
We introduced an incongruence-aware evaluation for multimodal SER in audio LLMs,
centred on a continuous score \pin{} validated against human judgement on all three
corpora, most thoroughly on eNTERFACE'05. Cross-modal incongruence independently predicts LLM error wherever
audio is present, an effect that is absent, and indeed reversed, in supervised
VAD-feature classifiers trained on the same features. We are
explicit about the strength of evidence: with the model held constant across
modalities, the \emph{established} result is the text+audio incongruence penalty,
present in all three corpora for GPT-audio (OR $1.15$--$1.25$) and shared in
direction by Qwen2.5-Omni on MELD; text-only associations are inconsistent in sign
and demonstrably model-dependent (Section~\ref{sec:modelabl}), so we draw no
conclusions from them.
The code, per-sample VAD features, \pin{} scores, validity-checked model
predictions and prompt templates are released as an incongruence-aware benchmark at
\url{https://github.com/Guylaban/incongruence-aware-ser}; the eNTERFACE'05, MELD and
CMU-MOSEI corpora themselves are available from their original distributors. We
recommend incongruence-stratified reporting as a standard robustness check for
multimodal SER.

\section*{Acknowledgment}
The authors declare no competing interests. As a non-native English speaker, the
first author used generative AI tools for spelling and grammatical correction; all
results were computed by the released code and verified by the authors, who take
full responsibility for the content.

\bibliographystyle{IEEEtranN}
\bibliography{mybib}

\begin{IEEEbiographynophoto}{Fengji Jing}
is with the Faculty of Science, the University of Hong Kong, Hong Kong, China.
\end{IEEEbiographynophoto}
\begin{IEEEbiographynophoto}{Guy Laban}
is with the Department of Industrial Engineering and Management, the School of Brain
Sciences and Cognition, and the Azrieli National Center for Autism and
Neurodevelopment Research, Ben-Gurion University of the Negev, Beer Sheva, Israel.
\end{IEEEbiographynophoto}

\end{document}